%% file: main.tex
\documentclass[letterpaper]{article} 
\usepackage{aaai2026}
\usepackage{times}  
\usepackage{helvet}  
\usepackage{courier}  
\usepackage[hyphens]{url}  
\usepackage{graphicx} 
\usepackage{placeins}
\usepackage{natbib}  
\usepackage{caption} 
\usepackage{algorithm}
\usepackage{algorithmic}
\usepackage{booktabs}

\usepackage{newfloat}
\usepackage{listings}
\DeclareCaptionStyle{ruled}{labelfont=normalfont,labelsep=colon,strut=off} 
\floatstyle{ruled}
\newfloat{listing}{tb}{lst}{}
\floatname{listing}{Listing}
\title{CountSteer: Steering Attention for Object Counting in Diffusion Models}

\author{
    Hyemin Boo\thanks{These authors contributed equally.} \quad
    Hyoryung Kim\footnotemark[1] \quad
    Myungjin Lee\footnotemark[1] \quad
    Seunghyeon Lee\footnotemark[1] \quad \\
    Jiyoung Lee \quad
    Jang-Hwan Choi \quad
    Hyunsoo Cho\thanks{Corresponding author.}
}
\affiliations{

    Ewha Womans University, Republic of Korea \\

    \texttt{\{hyeminb, michellekim0922, lmjin, 2277044, lee.jiyoung, choij, chohyunsoo\}@ewha.ac.kr}
}

\usepackage{bibentry}
\usepackage{amsmath}
\usepackage{graphicx}     
\usepackage{capt-of}   
\usepackage{microtype}
\usepackage{cuted}

\begin{document}
\maketitle

\input{CountSteer/Section/00-abs-HS}

\input{CountSteer/Section/01-int-HS}

\input{CountSteer/Section/02-pre}

\input{CountSteer/Section/03-met}

\input{CountSteer/Section/04-exp}

\input{CountSteer/Section/05-con}

\bibliography{aaai2026}

\input{CountSteer/Section/99-app}

\end{document}

%% file: CountSteer/Section/00-abs-HS.tex
\begin{strip}
    \vspace{-2.5cm}
    \centering
    \includegraphics[width=\textwidth]{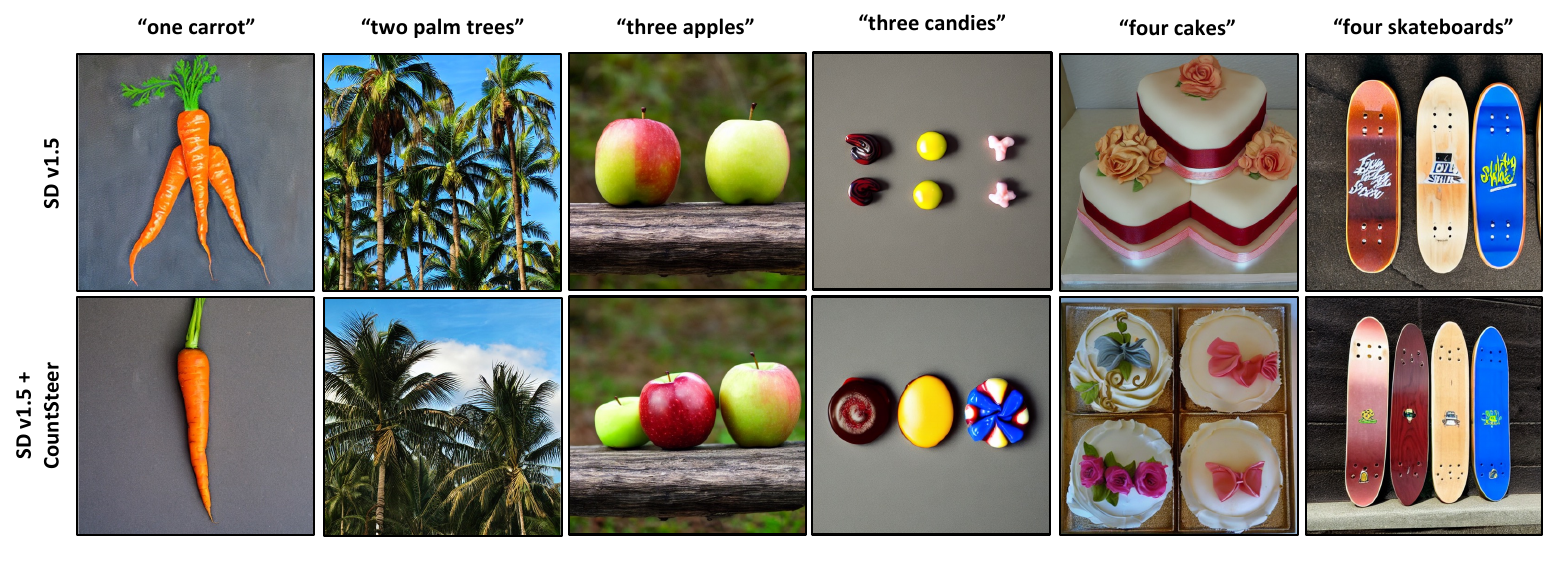}
    \vspace{-0.5cm}
    \footnotesize
    \captionof{figure}{
    \textbf{CountSteer improves numerical fidelity in text-to-image diffusion.}
    While baseline Stable Diffusion v1.5 often fails to generate the correct number of objects, our cross-attention steering framework generates images with more accurate object counts that better align with user-specified prompts across diverse object categories.
    }
    \label{fig:teaser}
    \vspace{0.3cm}
\end{strip}

\vspace{-10pt}

\begin{abstract}
Text-to-image diffusion models generate realistic and coherent images but often fail to follow numerical instructions in text, revealing a gap between language and visual representation.
Interestingly, we found that these models are not entirely blind to numbers—they are \textit{implicitly aware} of their own counting accuracy, as their internal signals shift in consistent ways depending on whether the output meets the specified count. 
This observation suggests that the model already encodes a latent notion of numerical correctness, which can be harnessed to guide generation more precisely.  
Building on this intuition, we introduce \textit{CountSteer}, a training-free method that improves generation of specified object counts by steering the model’s cross-attention hidden states during inference.
In our experiments, \textit{CountSteer} improved object-count accuracy by about 4\% without compromising visual quality, demonstrating a simple yet effective step toward more controllable and semantically reliable text-to-image generation.
\end{abstract}

%% file: CountSteer/Section/01-int-HS.tex
\begin{figure*}[t]
    \centering
    \includegraphics[width=0.8\textwidth]{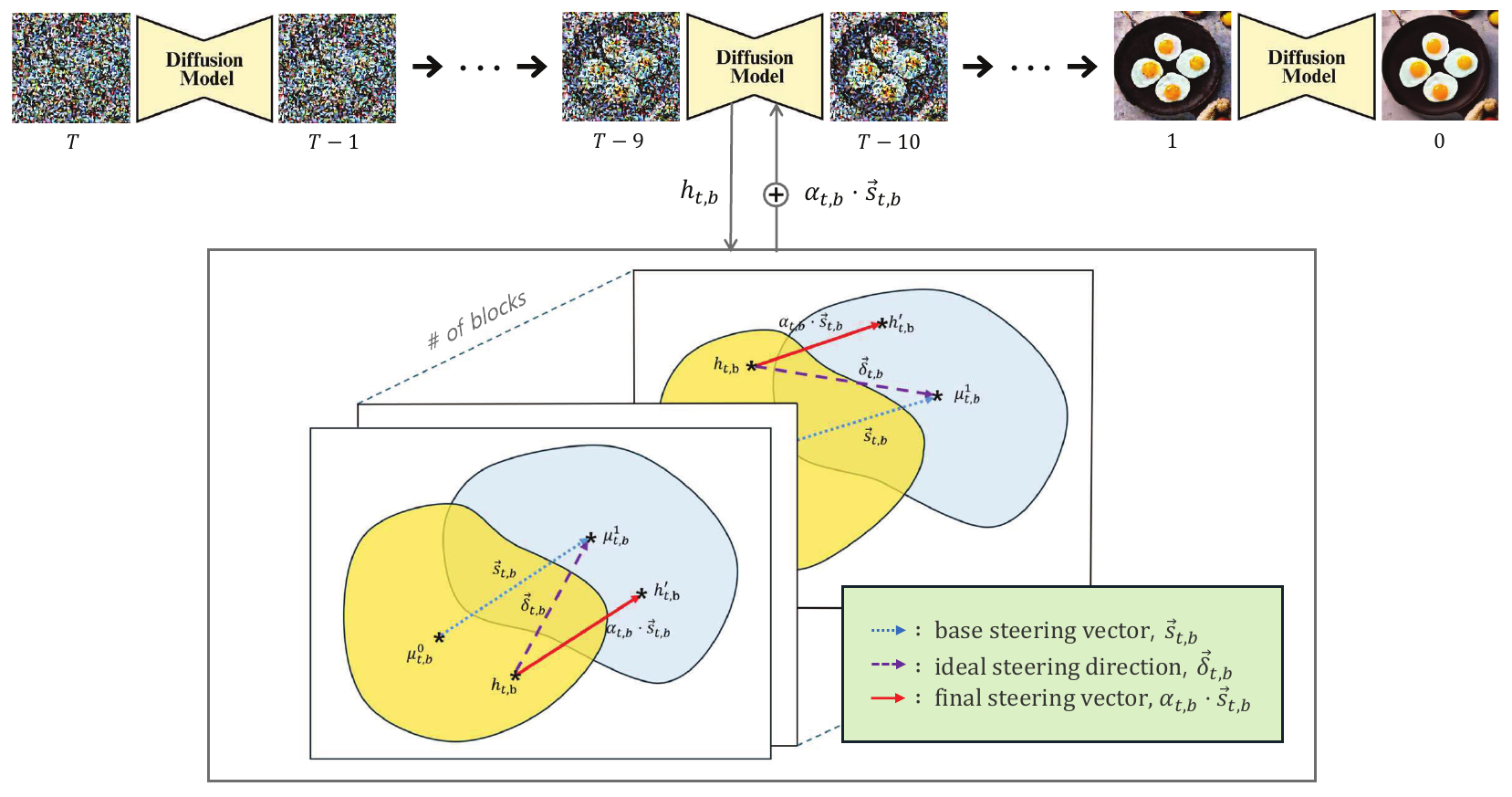}
    \caption{CountSteer: During each denoising step $t$, adaptive steering vectors are injected into UNet blocks to steer hidden states based on latent distributional differences between correct ($\mu^1_{t,b}$) and incorrect ($\mu^0_{t,b}$) samples. }
    \label{fig:architecture}
\end{figure*}

\vspace{-10pt}

\section{Introduction}
Recent progress in text-to-image (T2I) diffusion models such as Stable Diffusion~\cite{rombach2022high} has expanded the frontier of generative AI, enabling models to create realistic and semantically rich images from natural language descriptions.
These models have been applied across diverse domains, including creative design, simulation, education, and synthetic data generation for downstream AI training, serving as powerful tools that bridge text and visual expression through controllable and cost-efficient synthesis.

The effectiveness of diffusion models fundamentally depends on fidelity, which encompasses how faithfully generated images capture the intended semantics, structure, and quantitative details of the input text. 
While fidelity can be viewed through semantic, structural, and quantitative aspects, current models still face persistent challenges in maintaining fine-grained accuracy, particularly in object composition, color consistency, and object count~\cite{compbench, imagen}.
Among these factors, object count provides a clear and quantifiable measure of fidelity.
However, diffusion models often struggle to maintain numerical accuracy, as they must correctly interpret and preserve count information throughout a stochastic and spatially complex generation process.
This frequently results in under- or over-generation of objects, thereby reducing reliability in applications that demand semantic precision and quantitative accuracy.

To gain deeper insight into this limitation, we hypothesize that diffusion models possess an internal understanding of numerical concepts but fail to effectively utilize this knowledge during image generation.
We therefore examine how the model’s internal representations behave when it succeeds or fails to follow numerical instructions.
Our analysis reveals that the latent distributions of cross-attention representations are distinguishable, exhibiting directional patterns that implicitly encode numerical awareness.
These structured signals, which can be represented as a global vector in the attention space, indicate that diffusion models already contain latent cues related to counting behavior (see Section~\ref{subsec:KDE} for details).

Building on this intuition, we introduce \textit{CountSteer}, a training-free inference-time method that leverages these latent cues to enhance object-count fidelity. 
Unlike prior approaches that rely on fine-tuning~\cite{hu2022lora,ruiz2023dreambooth} or architectural modifications~\cite{huang2023composer,mou2024t2i}, our method operates directly during inference without retraining or altering the model structure.
Specifically, \textit{CountSteer} identifies a directional pattern within the model’s attention space that separates correct from incorrect generations and uses it to gently steer the diffusion trajectory toward numerically consistent outcomes.
By modulating this steering signal adaptively during inference, \textit{CountSteer} enhances count accuracy without altering model parameters or compromising image quality. 
Empirically, we observe an average 4\% improvement in object-count fidelity, demonstrating that small, geometry-based interventions can unlock latent reasoning capabilities in large generative models.

\begin{figure*}[t]
    \centering
    \includegraphics[width=\textwidth]{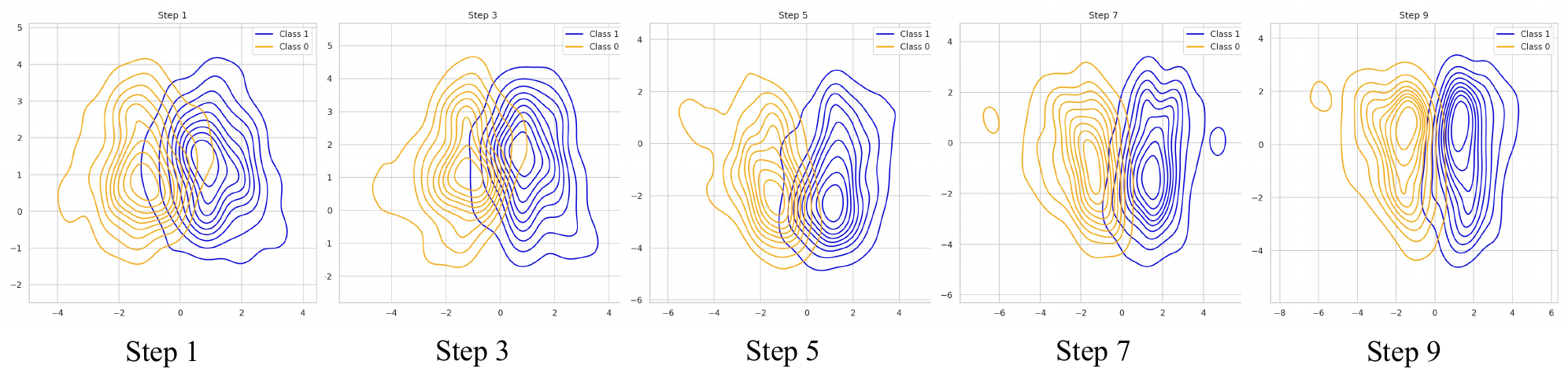}
    \caption{ Visualization of hidden state distributions analyzed via KDE. The two classes (correct and incorrect counts) exhibit clearly separable regions, supporting the formulation of our steering direction vector.}
    \vspace{-0.5cm}
    \label{fig:kde_analysis}
\end{figure*}

%% file: CountSteer/Section/02-pre.tex
\section{Preliminary}
\subsection{Steering}
Steering is a training-free method to control a model’s latent feature space. During generation, the model tends to form distinct hidden-state distributions for correct (Class 1) and incorrect (Class 0) samples. We vectorize this distributional difference and use it as a steering direction to shift the model’s representation toward the correct distribution. This phenomenon was also observed in Inference-Time Intervention ~\cite{li2023inference}, where hidden representations of language models exhibited linearly separable directions corresponding to truthful and untruthful classes. In ITI, interventions were applied to head-level activations of multi-head attention at each transformer layer. We adopt this concept to the structure of Stable Diffusion, performing interventions at each denoising step $t$ within the UNet blocks $b$. Furthermore, we empirically examine whether a similar separation emerges in a diffusion model between Class 1 and Class 0. 

For each class, we compute the mean of hidden state vectors for the Class 1 and Class 0 samples, denoted as $ \mu^1 $ and $ \mu^0 $. The difference  $ \vec{s} = \mu^1 - \mu^0$ is then defined as the steering direction vector, which guides the model toward improved output alignment:

\vspace{-2pt}

\begin{equation}
    \mu_1 = \frac{1}{N_1}\sum_{i=1}^{N_1} h_i^{(1)}, ~~
    \mu_0 = \frac{1}{N_0}\sum_{j=1}^{N_0} h_j^{(0)} \label{eq:mean} 
\end{equation}
\begin{equation}
    h_t' = h_t + \alpha \cdot \vec{s} \label{eq:steering}
\end{equation}

\vspace{-0.5pt}

During inference, the hidden state $h_t$ is adjusted by adding the steering direction $\vec{s}$. Here, $\alpha$ controls the strength of the adjustment. We further extend this formulation across diffusion steps, defining a step and block-wise steering vector as $\vec{s}_{t,b} = \mu^1_{t,b} - \mu^0_{t,b}$ and applying it dynamically throughout the denoising process.

\subsection{Numerical Steering via KDE Analysis}
\label{subsec:KDE}

We analyze the distributions in the image generation process with respect to object count fidelity. The samples are divided into Class 1 and Class 0. Visualization via Kernel Density Estimation (KDE) shows that the query hidden states of these two classes represent separable distributions. This observation suggests that, while the model inherently has the ability to represent object counts, such a capability is not consistently preserved during inference. Based on this observation, we hypothesize that steering the hidden representation of a generated sample to the distribution of Class 1 can improve count alignment. To achieve this, we define $ \vec{s}_{t,b} $ as the difference between the mean hidden states of Class 1 and Class 0, and apply it to adjust the latent representation during inference. This approach provides a training-free mechanism to improve numerical fidelity.


%% file: CountSteer/Section/03-met.tex
\section{Method}
Our analysis reveals that text-to-image diffusion models' hidden states form two distinct distributions based on object count correctness. While steering vectors can shift representations between these distributions, prompt-dependent variability causes fixed vectors to yield insufficient or excessive corrections. We therefore introduce two adaptive scaling factors that modulate the steering vector's magnitude and direction, enabling prompt-specific calibration.

\begin{figure}[t]
    \centering
    \includegraphics[width=\linewidth]{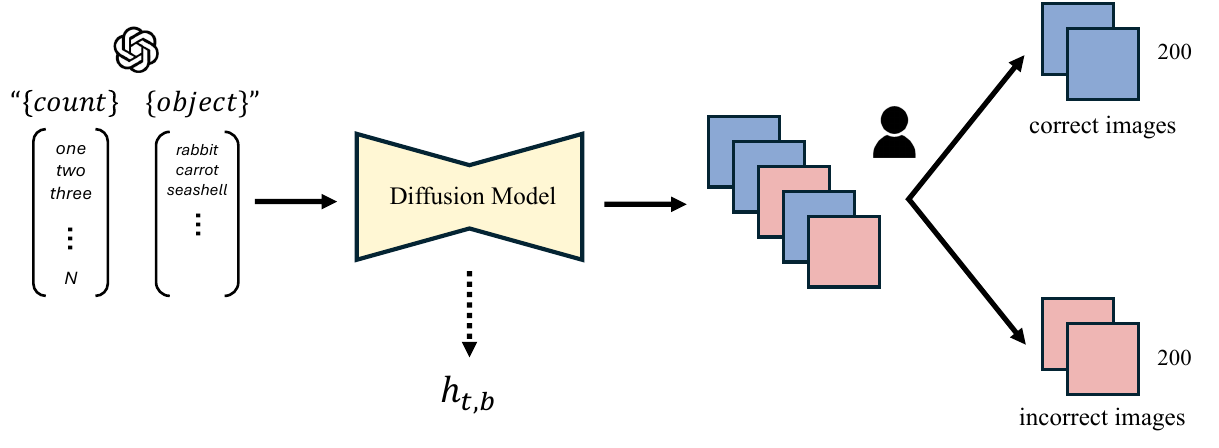}
    \vspace{-0.4cm}
    \caption{We generate images from "\{count\} \{object\}" prompts, extract hidden states $h_{t,b}$, and manually annotate them into balanced classes of 200 correct and incorrect counting to construct the base steering vector.}
    \vspace{-0.5cm}
    \label{fig:dataset}
\end{figure}

\subsection{Dataset Construction}
As shown in Figure~\ref{fig:barchart} of the Appendix, the model’s accuracy in producing the correct number of objects decreases as the target count increases. Beyond four, the generations almost entirely fail, suggesting that the model lacks consistent internal representations for higher quantities.
Therefore, our dataset focuses exclusively on the one-to-four range, where the model demonstrates a rudimentary yet analyzable understanding of numerical concepts. To construct the two distributions, we create a dataset of 600 prompts following the template \{count\} \{object\}” (e.g., “three cats”, “one carrot”), with equal numbers for each count from one to four. These prompts, automatically generated using GPT-4o, are partitioned into 400 for steering vector construction and 200 for evaluation, with no overlap between the two sets. For the steering vector construction dataset, we generate one image per prompt and manually annotate each as class 0 or class 1. To maintain class balance across all prompts, we regenerate images using different random seeds when needed until achieving equal representation of both classes. From these annotated images, we extract hidden states during the diffusion process—specifically, the query vectors from cross-attention layers at the first $k$ denoising timesteps across all UNet blocks.

\subsection{CountSteer}
Our approach addresses the limitation of fixed steering vectors by introducing an adaptive scaling mechanism. We dynamically adjust both the magnitude and direction of steering based on the current hidden state's position relative to the target distribution.

\begin{equation}
    s_{t,b}=\mu^1_{t,b} - \mu^0_{t,b}
    \label{eq:steering_vector} 
\end{equation}

We first construct a base steering vector $s_{t,b}$ that represents the average direction from incorrect to correct counting distributions at each denoising step $t$ and UNet block $b$. The $\mu^1_{t,b}$ and $\mu^0_{t,b}$ denote the mean hidden states of correct (class 1) and incorrect (class 0) distributions, respectively. 

\begin{equation}
    d_{t,b} = \frac{||\delta_{t,b}||_2}{||s_{t,b}||_2}, ~~ \delta_{t,b} = \mu^1_{t,b} - h_{t,b}
    \label{eq:adpative_scaling} 
\end{equation}

While $s_{t,b}$ provides the general steering direction, its effectiveness varies with varying prompts. To address this, we introduce an adaptive scaling factor $\alpha_{t,b}$ that adjusts the steering intensity based on two key measurements. First, we compute the distance-based scaling as $d_{t,b}$, where $\delta_{t,b}$ represents the vector from the current hidden state to the target distribution mean. This ratio indicates whether the current state requires stronger or weaker steering. Second, we assess the directional alignment through cosine similarity, which determines whether the base steering vector points in the appropriate direction — positive values maintain the original direction while negative values reverse it.

\begin{equation}
    \alpha_{t,b} = \cos(s_{t,b}, \delta_{t,b}) \cdot (1 - e^{-d_{t,b}}) \cdot c
    \label{eq:adpative_scaling} 
\end{equation}

The adaptive scaling factor combines both measurements. This formulation ensures smooth scaling: when the hidden state is far from the target ($d_{t,b} > 1$), the exponential term approaches 1, allowing stronger steering; conversely, when close to the target ($d_{t,b} < 1$), steering is reduced to prevent overshooting. The constant $c$ serves as a global scaling factor that amplifies the overall steering magnitude, as the product of similarity and the exponential term alone produces values too small for effective correction. We empirically set $c=100$ to control the scaling sensitivity. 

\begin{equation}
    h'_{t,b} = h_{t,b} + \alpha_{t,b} \cdot s_{t,b}
    \label{eq:updating_vector} 
\end{equation}

The adapted steering vector is then applied to update the hidden state. During inference, we inject these adaptive steering vectors at each denoising step, enabling prompt-specific corrections. We evaluate this approach on 100 test images generated with the same prompt template used for training.

%% file: CountSteer/Section/04-exp.tex
\section{Experiment}

We adopt Stable Diffusion v1.5 as the backbone model, performing inference with 50 denoising steps, and a guidance scale of 7.5. 
Following prior studies ~\cite{latentspace, layoutcontrol, painters} suggesting that global layout and coarse structure are primarily determined during the early denoising stages, the proposed steering vector is applied during the first 10 denoising steps ($k = 10$).
This early-stage intervention empirically provides a good balance between controllability and image fidelity. 
All experiments are conducted with randomly initialized seeds.

\subsection{Evaluation Protocol}
\begin{table}[!t]
\renewcommand{\arraystretch}{1.1}
\setlength{\tabcolsep}{6pt} 
\centering
\vspace{-0.5em}
\resizebox{\linewidth}{!}{ 
\begin{tabular}{l|cccc}
\toprule
\textbf{Methods} & \textbf{ACC} ↑ & \textbf{MAE} ↓ & \textbf{CLIP-Score} ↑ \\
\midrule
SD v1.5 (Baseline) & 50.0\% & 1.125 & \textbf{30.99} \\
\textbf{SD v1.5 + CountSteer} & \textbf{54.0\%} & \textbf{0.940} & 30.39 \\
\bottomrule
\end{tabular}
}
\vspace{-0.3em}
\caption{Quantitative comparison of Accuracy, Mean Absolute Error, and CLIP-Score.}
\label{tab:quantitative_all}
\end{table}
For quantitative evaluation, we employ the LLaVA-OneVision ~\cite{li2024llava} model to automatically count the number of objects in the generated images. 
The following instruction prompt is used for all evaluations: \textit{“How many \{objects\} are in the image? Reply with only a number.”} 
This prompt formulation ensures consistent and deterministic numeric responses across diverse object categories, allowing fair comparison between methods.

We evaluate performance using three metrics—Accuracy (ACC), Mean Absolute Error (MAE), and CLIP-Score. 
ACC measures whether the model produces the exact object count specified in the prompt, capturing counting fidelity. 
MAE reflects the average deviation between generated and target counts. 
The CLIP-Score ~\cite{taited2023CLIPScore} quantifies semantic alignment between the generated image and the text prompt, verifying that steering preserves intended meaning and visual coherence. Specifically, it measures the cosine similarity between the image and text embeddings produced by a pretrained CLIP ~\cite{radford2021learning} model, serving as an indicator of how well visual content corresponds to the textual description.

\subsection{Quantitative Results}
Table ~\ref{tab:quantitative_all} shows the proposed CountSteer framework consistently outperforms the baseline across all evaluation metrics. 
Specifically, CountSteer achieves a 4.0\%p increase in ACC, suggesting that the steering vector effectively helps the denoising trajectory toward the more accurate semantic and quantitative outcome. 
In terms of numerical deviation, the MAE is reduced by 0.185, indicating that CountSteer significantly mitigates extreme cases in which the model either under- or over-generates objects. 
Importantly, despite the additional intervention at inference time, the CLIP-Score remains comparable to that of the baseline. 
This finding confirms that CountSteer preserves semantic alignment between the generated image and the conditioning text, ensuring that improvements in quantitative fidelity do not come at the expense of semantic integrity or image quality.

\subsection{Qualitative Results}
Figure ~\ref{fig:teaser} demonstrates that CountSteer systematically improves object count accuracy while preserving semantic coherence and photorealistic quality. 
The baseline often suffers from under- or over-generation, whereas CountSteer adaptively increases or suppresses object production to correct such deviations, effectively achieving bidirectional control over object quantities. 
Furthermore, the generated images preserve fine-grained details, indicating that our steering mechanism selectively modulates counting-relevant representations without compromising other generative attributes or altering the overall visual style.

Collectively, these results substantiate our central hypothesis that diffusion models inherently encode the correct semantic and quantitative representations within their internal feature space but often fail to manifest them explicitly in the final outputs. 
By steering the model’s internal representations toward a semantically consistent direction, CountSteer effectively elicits this latent capability—enhancing counting accuracy and overall generation fidelity without any additional fine-tuning or retraining.

%% file: CountSteer/Section/05-con.tex
\section{Conclusion}


In this work, we introduced \textit{CountSteer}, an inference-time steering approach for improving object count fidelity in diffusion-based generative models. While foundation models such as Stable Diffusion often fail to align with prompt-specified object quantities, \textit{CountSteer} rectifies this limitation by manipulating latent space distributions without additional training. Our method computes a steering vector from the mean hidden-state discrepancy between correctly and incorrectly generated samples, dynamically scaled to each prompt via adaptive factors. This lightweight intervention improves counting accuracy by 4\% while preserving CLIP-Score and visual quality. Beyond object counting, our steering framework can also be applied to compositional attributes such as color, spatial arrangement, and multi-object interactions to further advance controllable and reliable synthetic data generation.

\section{Acknowledgments}
This research was supported by the High-Performance Computing Support Project, funded by the Government of the Republic of Korea (Ministry of Science and ICT).

%% file: CountSteer/Section/99-app.tex
\onecolumn
\appendix
\section*{Appendix}
\vspace{4pt}
\section{Distribution of Correct and Incorrect Object Counts}
\vspace{3pt}

\begin{figure}[H]
    \centering
    \includegraphics[width=0.7\linewidth]{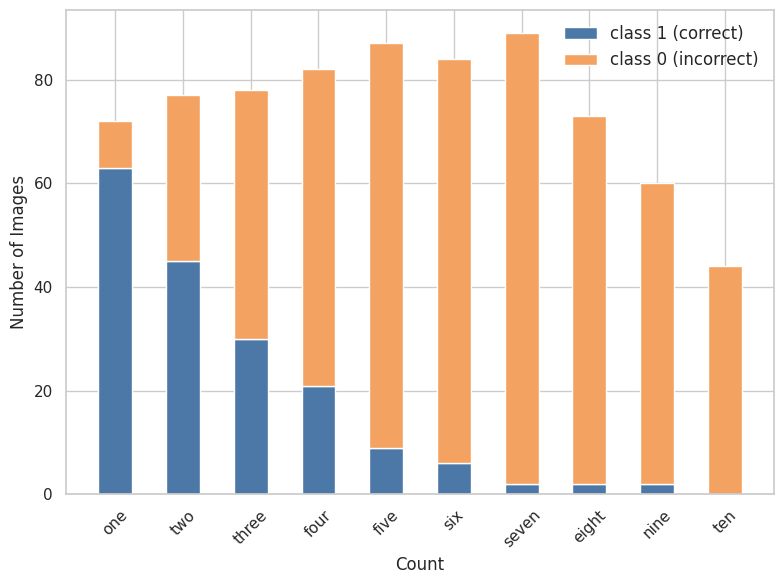}
    \caption{Overall distribution of correct and incorrect counts across different targets (one to ten).}
    \vspace{-0.5cm}
    \label{fig:barchart}
\end{figure}

Figure~\ref{fig:barchart} illustrates the overall distribution of correct and incorrect generations across different target counts.
As the specified number of objects increases, the model’s accuracy drops sharply, with nearly complete failure beyond a count of four.
This pattern indicates that Stable Diffusion v1.5 exhibits a limited and inconsistent internal representation of numerical concepts at higher quantities.
Therefore, the primary dataset used in our analysis focuses on the one-to-four range, where the model retains a basic yet interpretable understanding of object count.

\vspace{5pt}

\clearpage

\section{Object Count Formation in Early Denoising}
\begin{figure}[H]
  \centering
  \includegraphics[width=\textwidth]{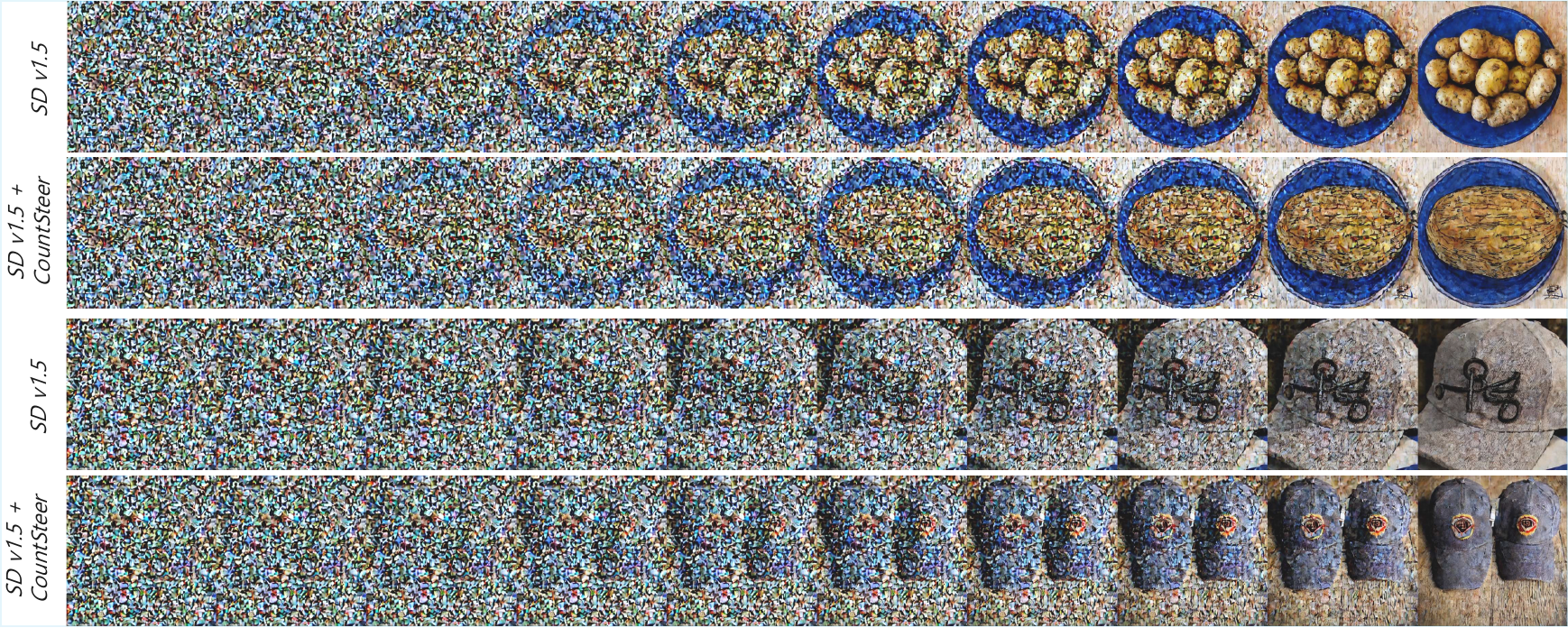}
  \vspace{-4pt}
  \includegraphics[width=\textwidth]{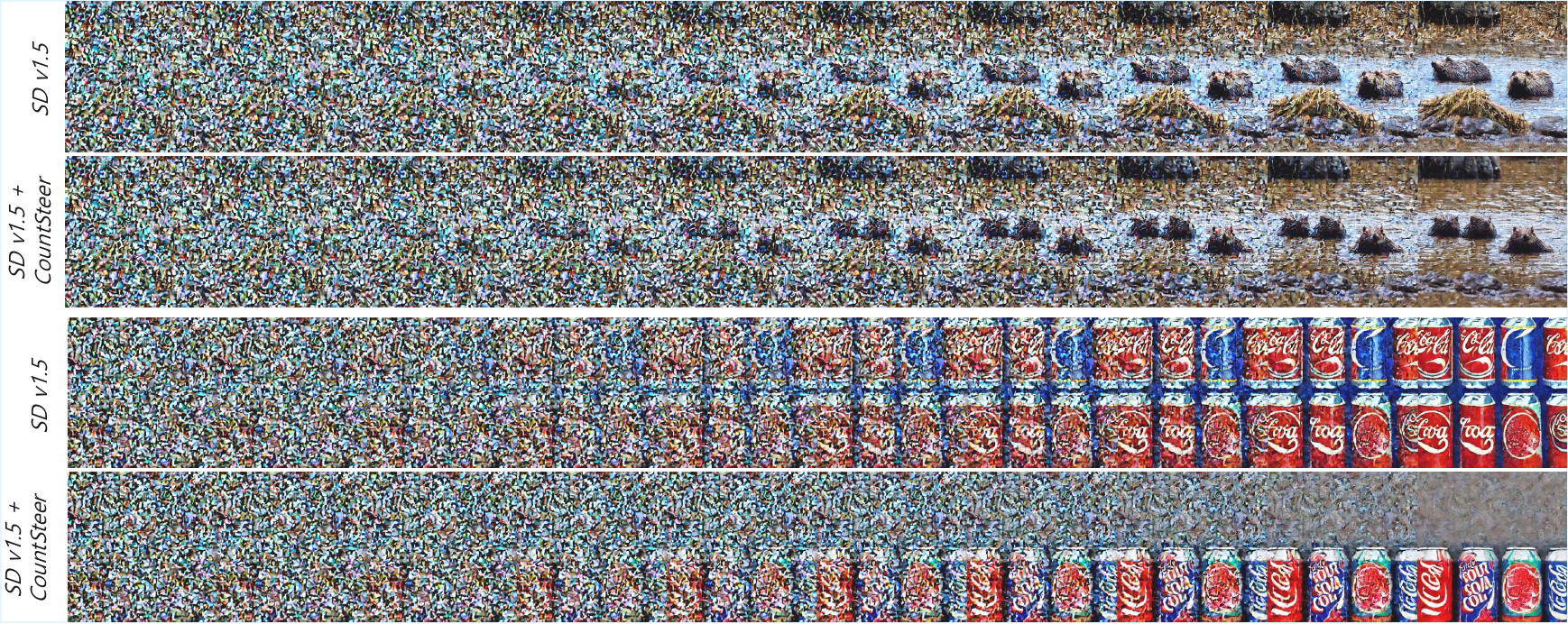}
  \vspace{-4pt}
  \caption{Visualization of denoising process. For each prompt, we visualized the denoising process for \textit{SD v1.5} and \textit{SD v1.5+CountSteer}. The prompts used to generate the images are, in order, \textit{'one potato', 'two baseball caps', 'three beavers', and 'four soda cans'}. }
  \label{fig:denoising_process}
\end{figure}

\vspace{3pt}
As shown in Figure \ref{fig:denoising_process}, we visualize the denoising processes of Stable Diffusion v1.5 and SD v1.5 + CountSteer using four representative prompts — \textit{“one potato”, “two baseball caps”, “three beavers” and “four soda cans”}. The visualization shows that the overall layout of each object tends to be determined in the early stages of the diffusion process. In particular, the approximate spatial arrangement and object count begin to emerge within the first few denoising steps. Based on this observation, we applied CountSteer only during the initial denoising phase (steps 1–10), and found that this early intervention effectively guided the model to form the desired object count and spatial configuration.

\vspace{5pt}

\onecolumn
\section{Additional Qualitative Results}
\vspace{5pt}
We present additional qualitative results comparing the baseline Stable Diffusion v1.5 and our CountSteer method across varying count values and object types. These results show that CountSteer generates images with object counts that align more accurately with the numerical prompts compared to the baseline model. 

\begin{figure*}[h]
    \centering
    \includegraphics[width=0.9\textwidth]{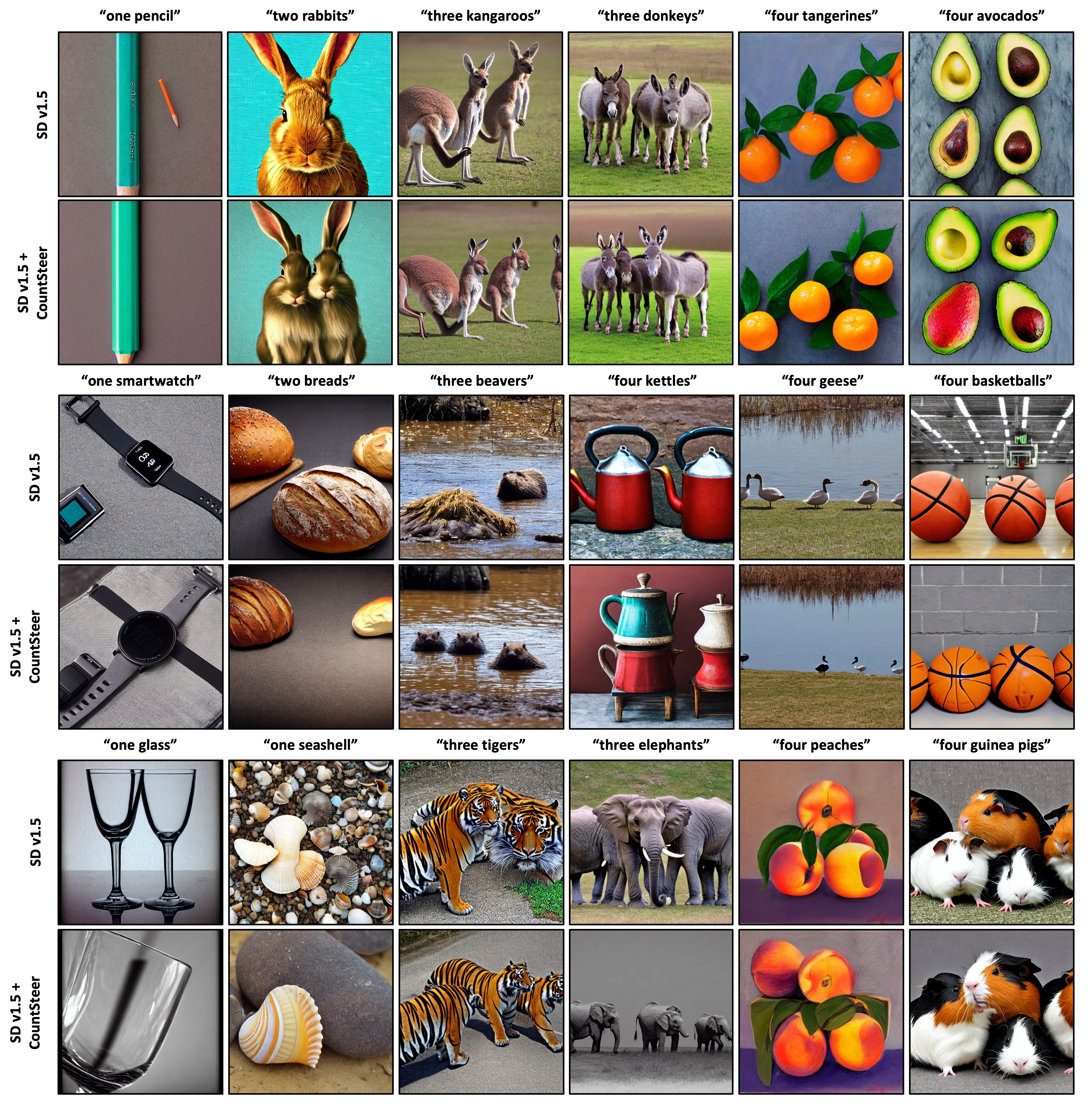}
    \caption{Additional examples showing the effect of CountSteer across various prompts.Compared to baseline Stable Diffusion v1.5, our method more consistently generates images with object counts that match the specified number across a wide range of object categories. }
    \vspace{-0.5cm}
    \label{fig:appendix_results}
\end{figure*}
\twocolumn

\onecolumn
\section{Failure Cases}

\begin{figure*}[h]
    \centering
    \includegraphics[width=\textwidth]{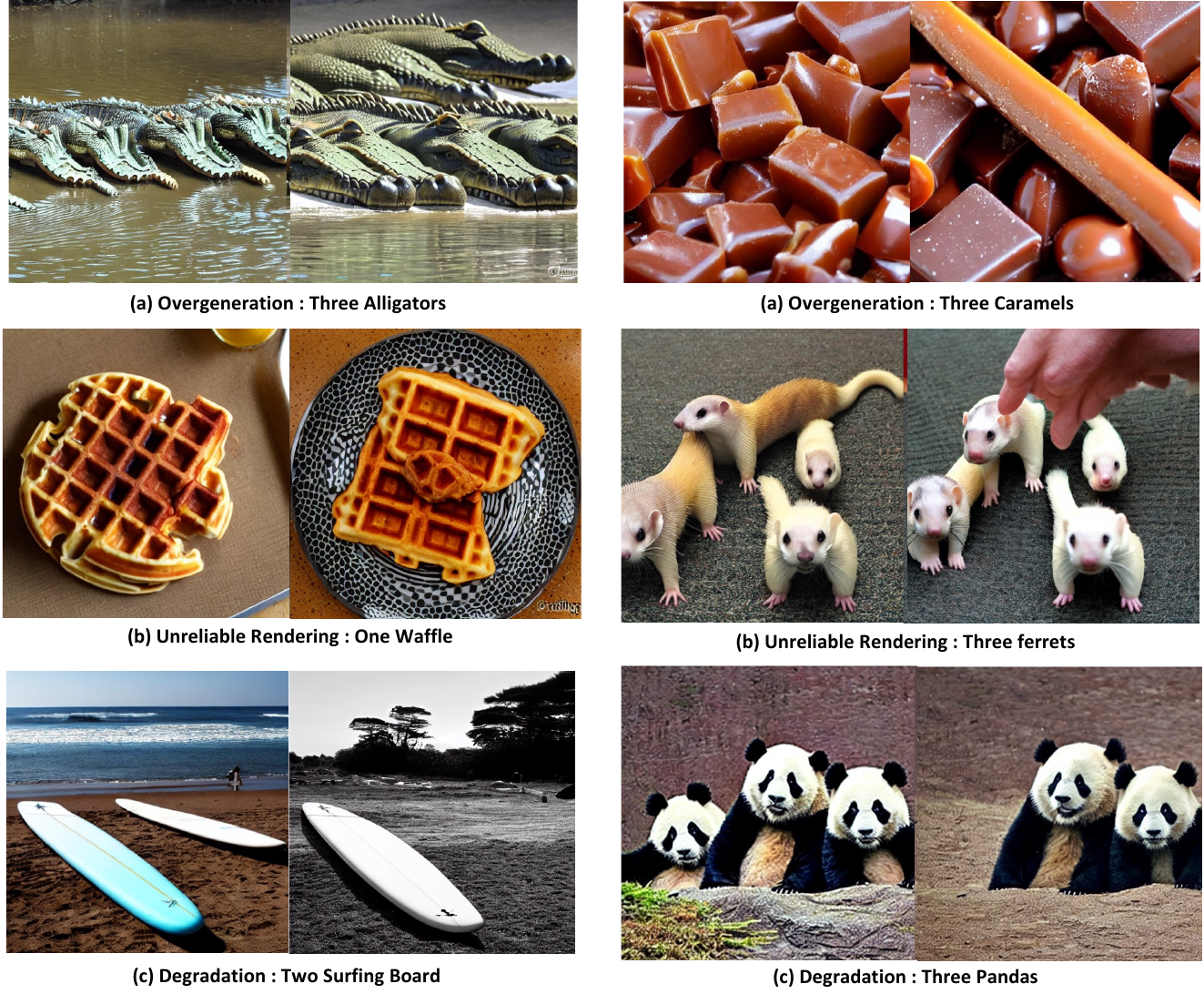}
    \caption{{Representative failure cases.} Each case highlights a different type of limitation, including overgeneration, rendering failure, object distortion, and regression from correct to incorrect output.}
    \label{fig:failure_cases}
\end{figure*}

Our analysis reveals three primary failure cases of the proposed steering method in Figure~\ref{fig:failure_cases}.
First, a phenomenon of over-generation occurs when the model produces more objects than intended, even under steering toward a lower count.
This behavior arises when class-wise latent directions excessively amplify object-related features, leading to duplication—particularly when the target object lacks a compact or well-defined representation in the latent space.
Second, we observe cases of unreliable rendering, where steering fails to take effect because the model inherently struggles to generate the target object.
If the object is rendered ambiguously or inconsistently regardless of the prompt, the steering signal derived from latent statistics does not correspond to meaningful semantic directions, revealing a dependence on the model’s baseline generative capacity.
Finally, the method may cause the degradation of initially correct output, where an originally accurate generation becomes incorrect after steering is applied.
This regression effect suggests the risk of over-steering, particularly when steering is applied uniformly without considering the confidence or quality of the base output.
These observations highlight the need for adaptive or selective steering mechanisms that dynamically respond to the model’s intermediate state.

\vspace{0.5cm}